\documentclass[preprint, journal]{vgtc}      

\graphicspath{{figures/}{pictures/}{images/}{./}} 

\usepackage{times}                     

\usepackage{mathptmx}                  
\usepackage{amssymb}
\usepackage{amsmath}
\onlineid{1842}

\vgtccategory{Research}

\vgtcpapertype{system}

\usepackage{booktabs}
\usepackage{graphicx}
\usepackage{enumitem}
\usepackage{soul}
\usepackage{dsfont}
\usepackage{gensymb}
\usepackage{microtype}
\usepackage{pifont}
\usepackage{threeparttable}
\usepackage{collect}
\usepackage{xspace}
\usepackage{xfrac}
\usepackage{array}
\usepackage{multirow}
\usepackage{wrapfig}
\usepackage{url}
\usepackage[nolist]{acronym}
\usepackage{siunitx}

\usepackage{pifont}
\newcommand{\cmark}{\checkmark}%
\newcommand{\xmark}{\scalebox{0.85}{\ding{53}}}%

\DeclareGraphicsExtensions{.png,.jpg,.pdf,.ai,.psd}
\def\figurePath{figures/}

\usepackage{xr}
\makeatletter
\newcommand*{\addFileDependency}[1]{
  \typeout{(#1)}
  \@addtofilelist{#1}
  \IfFileExists{#1}{}{\typeout{No file #1.}}
}
\makeatother

\newcommand{\refFig}[1]{Fig.~\ref{fig:#1}}

\newcommand{\refTab}[1]{Tab.~\ref{tab:#1}}

\newcommand{\refSec}[1]{Sec.~\ref{sec:#1}}

\usepackage[bold=0.55]{xfakebold}

\usepackage{icomma}

\newcommand{\todo}[1]{\textcolor{green}{TODO: #1}}

\newcommand{\myfigure}[2]{%
    \begin{figure}[htb]%
    \centering\includegraphics*[width = \linewidth]{\figurePath#1}%
    \vspace{-.1cm}%
    \caption{#2}%
    \vspace{-.2cm}%
    \label{fig:#1}%
    \end{figure}%
}

\newcommand{\mycfigure}[2]{%
    \begin{figure*}[htb]%
    \centering\includegraphics*[width = \linewidth]{\figurePath#1}%
    \vspace{-.1cm}%
    \caption{#2}%
    \vspace{-.2cm}%
    \label{fig:#1}%
    \end{figure*}%
}

\soulregister\ref7
\soulregister\cite7
\soulregister\refFig7
\soulregister\refAlg7
\soulregister\cite7
\soulregister\ref7
\soulregister\pageref7
\soulregister\shortcite7
\soulregister\eg0
\soulregister\ie0
\soulregister\etal0

\DeclareGraphicsExtensions{.png,.jpg,.pdf,.ai,.psd}
\newcommand{\mysection}[2]{\section{#1}\label{sec:#2}}
\newcommand{\mysubsection}[2]{\subsection{#1}\label{sec:#2}}

\newcommand{\eg}{e.g.,\ }
\newcommand{\ie}{i.e.,\ }
\newcommand{\etal}{et~al.\ }

\newcommand{\mymath}[2]{
    \newcommand{#1}{\TextOrMath{$#2$\xspace}{#2}}
}

\mymath{\inputImage}{I}
\mymath{\noDoFImage}{I_\mathrm{noDoF}}
\mymath{\noMBImage}{I_\mathrm{noMB}}
\mymath{\noNoiseImage}{I_\mathrm{noNoise}}
\mymath{\mbRemoval}{\mathtt{removeMB}}
\mymath{\dofRemoval}{\mathtt{removeDoF}}
\mymath{\noiseRemoval}{\mathtt{removeNoise}}
\mymath{\mbSynthesis}{\mathtt{addMB}}
\mymath{\dofSynthesis}{\mathtt{addDoF}}
\mymath{\noiseSynthesis}{\mathtt{addNoise}}
\mymath{\mbParams}{\lambda}
\mymath{\dofParams}{\delta}
\mymath{\noiseParams}{\sigma}
\mymath{\depthAnalysis}{\mathtt{getDepth}}
\mymath{\flowAnalysis}{\mathtt{getFlow}}
\mymath{\depthImage}{Z}
\mymath{\flowImage}{F}
\mymath{\dofImage}{I_\mathrm{DoF}}
\mymath{\mbImage}{I_\mathrm{MB}}
\mymath{\noiseImage}{I_\mathrm{Noise}}
\mymath{\loss}{\mathcal{L}}
\mymath{\dofLoss}{\loss_\mathrm{DoF}}
\mymath{\mbLoss}{\loss_\mathrm{MB}}
\mymath{\noiseLoss}{\loss_\mathrm{Noise}}

\preprinttext{To appear in IEEE Transactions on Visualization and Computer Graphics.}

\ieeedoi{xx.xxxx/TVCG.201x.xxxxxxx}

\begin{acronym}
\acro{MB}{motion blur}
\acro{MR}{mixed reality}
\acro{DoF}{depth of field}
\acro{MLE}{maximum-likelihood estimation}
\acro{PSF}{point spread function}
\acro{std}{standard deviation}
\end{acronym}

\newcommand{\distortions}{noise, \ac{MB} and \ac{DoF}\xspace}

\newcommand{\SP}[1]{\textcolor{blue}{\textbf{SP: #1}}}
\newcommand{\new}[1]{\textcolor{red}{#1}}
\renewcommand{\todo}[1]{ }
\renewcommand{\SP}[1]{ }
\renewcommand{\new}[1]{#1}

\newcommand{\add}[1]{\textcolor{Blue}{#1}}
\newcommand{\remove}[1]{\textcolor{Red}{\st{#1}}}
\renewcommand{\add}[1]{#1}
\renewcommand{\remove}[1]{}

\title{Blind Augmentation: Calibration-free Camera Distortion Model Estimation for Real-time Mixed-reality Consistency}

\author{%
  \authororcid{Siddhant Prakash}{0009-0000-8686-2442},
  \authororcid{David R. Walton}{0000-0001-5879-9714},
  \authororcid{Rafael K. dos Anjos}{0000-0002-2616-7541},
  \authororcid{Anthony Steed}{0000-0001-9034-3020}, and
  \authororcid{Tobias Ritschel}{0009-0006-4660-7790}
}

\authorfooter{
  \item
  	Siddhant Prakash, Anthony Steed, and Tobias Ritschel are with University College London.
  	E-mail: \{s.prakash\,$|$\,a.steed\,$|$\,t.ritschel\}@ucl.ac.uk\,.
  \item
  	David R. Walton is with Birmingham City University.
  	E-mail: david.walton@bcu.ac.uk\,.
  \item
  	Rafael K. dos Anjos is with University of Leeds.
  	E-mail: r.kuffnerdosanjos@leeds.ac.uk\,.
}

\teaser{
  \centering
  \includegraphics[width=\linewidth]{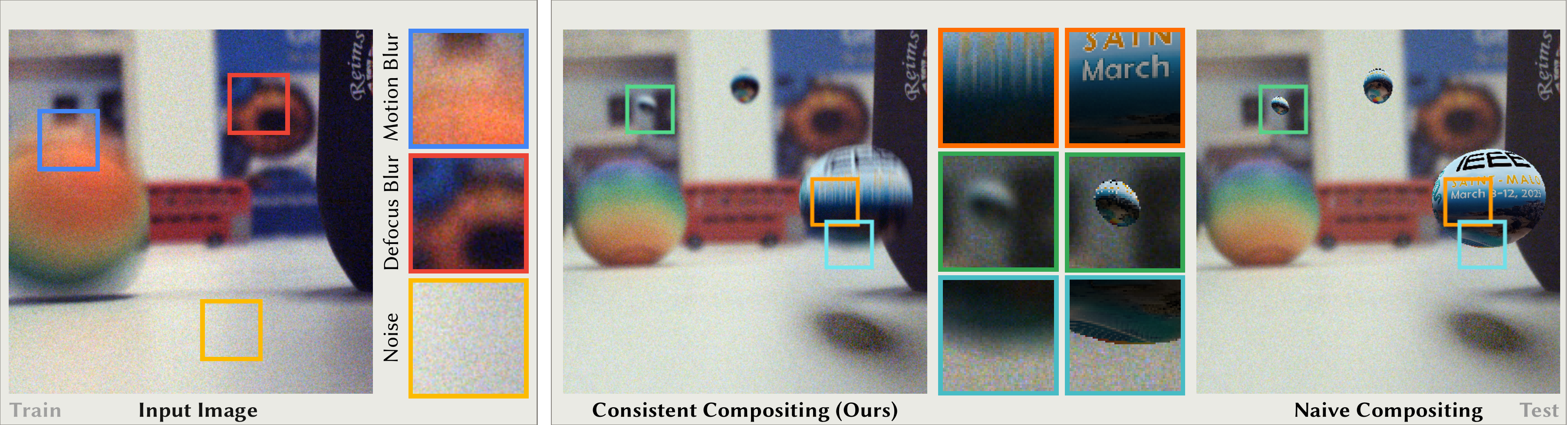}
  \caption{Our method ``blindly'' estimates a model of \distortions from input frames (left), \ie without requiring any known calibration markers / objects.
    The model can then augment other images with virtual objects that appear \emph{visually consistent} (middle).}
  \label{fig:teaser}
}

\abstract{
Real camera footage is subject to noise, motion blur (MB) and depth of field (\remove{Dof}\add{DoF}).
In some applications these might be considered distortions to be removed, but in others it is important to model them because it would be ineffective, or interfere with an aesthetic choice, to simply remove them. 
In augmented reality applications where virtual content is composed into a live video feed, we can model \distortions to make the virtual content visually consistent with the video.
Existing methods for this typically suffer two main limitations.
First, they require a camera calibration step to relate a known calibration target to the specific cameras response.
Second, existing work require methods that can be (differentiably) tuned to the calibration, such as slow and specialized neural networks.
We propose a method which estimates parameters for \distortions instantly, which allows using off-the-shelf real-time simulation methods from \eg a game engine in compositing augmented content.
Our main idea is to unlock both features by showing how to use modern computer vision methods that can remove \distortions from the video stream, essentially providing self-calibration.
This allows to auto-tune any black-box real-time noise+\ac{MB}+\ac{DoF} method to deliver fast and high-fidelity augmentation consistency.
} 


\begin{document}

\firstsection{Introduction}
\label{sec:Introduction}

\maketitle

\new{Video passthrough \ac{MR}} involves combining captured images and \new{computer-generated} images.
Captured images come with many artifacts of the capture process such as motion blur (MB), depth of field (DoF), vignetting, noise, etc.
While \new{in recent mixed-reality headsets,} capture is often set up so that artifacts are minimized, in many situations, they cannot be removed.
Indeed captured images have characteristic aesthetic qualities that we might want to reproduce \new{as is evident from the fact that many modern game engines provide methods to artificially introduce such artifacts in production footage for VFX}.
\new{Even where possible, removing all such artifacts may be undesirable, and limit creative expression.}
Thus \new{our} goal is to make the \new{computer-generated} images \emph{visually consistent} with the captured images by \new{reproducing}
some of these artifacts. 

Real-time \ac{MR} presents specific challenges within this larger problem.
Convincing \ac{MR} systems model several characteristics \cite{azuma1997survey}: scene geometry, materials, illumination and parameters characteristics of the capture camera.
\ac{MR} systems run at high framerates, might use multiple cameras that have a relatively small form factor, and those cameras are attached to moving people or devices.
Thus when processing video for \ac{MR} systems, the system must deal \new{with}
images with large amounts of \distortions. 

Typically the solution is to calibrate the virtual camera, manually \cite{klein2008compositing,fischer2006enhanced,aittala2010inverse,park2009esm} or by means of clean-distorted image pairs \cite{okumura2006augmented} eventually used as training data for learning \cite{mandl2021neural}.
Each of these methods requires time and/or expert knowledge, requires markers and/or limits camera settings.
Moreover, commercial video see-through augmented reality on mobile phones or head-mounted displays (HMD) may apply closed-box post-processing algorithms to the images, using unknown parameters. 
An ideal \ac{MR} system would not require calibration and could deal with dynamic changes in focus, noise, lighting levels, etc. 

We propose \emph{blind augmentation} (\refFig{teaser}), which can achieve visual consistency between captured and synthesized images, but without active calibration.
It is ``blind'' in the sense of ``blind deconvolution'' \cite{kundur1996blind, levin2009understanding}, which can reconstruct a distortion (\eg a convolution, \ie a blur) without knowing the precise nature of the particular distortion.
The twist in our work is that we are not seeking to remove distortion from real videos, but to blindly introduce distortion to virtual content to make it consistent with the captured images.
The key to make this work is recent methods to extract depth and motion as well as to remove \distortions from video streams. Our contributions are:
\begin{itemize}
    \item A method to estimate parameters for \distortions from a wild video without the need of calibration patterns, or camera metadata/hardware information. 
    \item Parameter update with an upper bound of \qty{1.5} sec., which allows video see-through augmented reality to be composited with \emph{consistent} quality shortly after start-up.
    \item A demo application demonstrating effectiveness of our method on the Meta~Quest~3 HMD running at \qty{45}\Hz.
\end{itemize}

\remove{We will release all code and datasets used in the paper.}
\add{All code and datasets used in the paper can be found on the project page: \url{https://prakashsidd18.github.io/projects/blind_augmentation/}.}


\mysection{Previous work}{PreviousWork}

Our work tackles visual consistency between live video and an augmented virtual rendering as required by real-time \ac{MR} systems \cite{bimber2005spatial,schmalstieg2016augmented,collins2017visual}.
Certainly, the view pose, geometry, reflectance and light need to be modeled or estimated, as pioneered by Fournier et al. \cite{fournier1993common} and reviewed by  Alhakamy and Tuceryan \cite{alhakamy2020real}.
In our work we assume these are known.
Instead, we look into the specific problem of recreating distortions that are found in a live video stream.

Klein and Murray \cite{klein2009simulating}, Fischer et al. \cite{fischer2006enhanced}, Aittala \cite{aittala2010inverse} and Park et al. \cite{park2009esm} simulate a number of image degradation effects for a particular small wide-angle camera, including barrel distortion, color response, chromatic aberration and vignetting.
Motion blur due to camera rotation (but not translation) is also simulated.

An early work in this area, Okumura et al. \cite{okumura2006augmented}, fits a \ac{PSF} to a real image based on the appearance of a circular AR marker placed in the image. 
This is then convolved with virtual content to add motion blur and \ac{DoF} effects. 
Its main limitations are the need for a known AR marker to be present in the scene, and estimation of a single global \ac{PSF}.
This means the approach can only handle static virtual objects at a similar depth to the marker.
We handle dynamic virtual objects at any depth, without the need for known AR markers in the scene.

The most strongly related work to our own is Mandl et al. \cite{mandl2021neural}.
The authors estimate camera parameters to introduce depth of field blur, vignetting and sensor artifacts such as noise.
They fit a complex camera model through capturing calibration images of a priori known markers and color charts.
From the data, a number of neural networks are trained, specific to the particular camera.
In contrast, our method requires no prior camera-specific training and can be used to augment arbitrary real-world videos.
\add{A follow-up work by the same authors~\cite{mandl24neuralbokeh} introduces neural bokeh which learns to synthesize high-quality out-of-focus effects for MR applications.}
We additionally handle motion blur, which is critical to generating consistent fast-moving augmented video sequences.
More advanced \ac{DoF} models can be used \cite{kan2012physically} and markers can be replaced by other patterns, \cite{pilet2006all}, but still these need to be known.

In a related work in a slightly different application area, Brostow and Essa \cite{brostow2001image} use computer vision techniques to add motion blur to stop-motion animated sequences, demonstrating the enhancement that consistent motion blur can make to perceived motion.

We note an alternative tactic where the real video stream can be made closer to the rendering one, by using  stylized rendering \cite{fischer2005stylized,steptoe2014presence} instead of trying to make the rendering more realistic.
We focus on making the augmentations more realistic, but this is an interesting direction for future work. 

Finally, methods such as (Deep) Image Harmonization aim to combine images in a harmonic \ie consistent way \cite{sunkavalli2010multi,xue2012understanding,tsai2017deep}.
By contrast we deal with virtual content where radiometrics and other differences are resolved by refined simulation, and the distortion model is fixed and somewhat known, but what is missing and changing between deployments is the camera distortion parameters. 

\mycfigure{Overview}{
Overview of our approach, comprising of a training part \emph{(top half)} and a test or deployment phase \emph{(bottom half)}.
Training starts with the input image \inputImage at the top left that is fed into off-the-shelf depth and flow extractors, as well as off-the-shelf methods to remove \distortions.
These off-the-shelf processes are denoted as (\emph{black arrows}).
Next, the image difference between a re-synthesis of \distortions is computed and compared to the input image (\emph{orange arrows}).
This error is minimized by back-propagating to the \distortions parameters (\emph{blue arrows}).
This forms a model that knows the noise profile, how to blur for which depth or which motion (\emph{top right}).
At test time, we know flow and depth of a virtual RGB image, and hence can synthesize \distortions (\emph{pink arrows}) using off-the-shelf and fast methods, before composing a final image with consistency superior to no \distortions.
}

\mysection{Our Approach}{OurApproach}
An overview of our approach is shown in \refFig{Overview}.
It has two main parts. 
The first is training (\refSec{Training}), that can be performed on any image captured from the camera system in question without the need for a known calibration and within seconds.
This outputs a model relating \distortions to scene content.
The second is testing (\refSec{Testing}), where a real-time rendering engine uses the \distortions model to add these effects.

\mysubsection{Training}{Training}
Training involves removing \distortions from an input and then re-inserting it subject to the right parameters, depending on the depth and motion maps extracted from the input as well, such that it matches the input.
We will detail each step next:

\paragraph{Input}
Training can be performed on any set of images \new{which exhibit \distortions}.
The \new{camera} does not need to see a known calibration target, but a natural scene such as a table-top, a car interior or a street.
For simplicity, we will explain the process for a single input image \inputImage.
If multiple images are to be used, a final model \new{could} be a suitable (robust) aggregate of the models for each individual image.

\begin{table*}[]
    \centering
    \caption{Operations involved, and methods used.}
    \label{tab:Ops}
    \begin{tabular}{llllllll}
    \toprule
    \multicolumn1c{Real}&
    \multicolumn1c{Virt.}&
    \multicolumn2c{Input Image}&
    \multicolumn2c{Output}&
    \multicolumn1c{Operator}&
    \multicolumn1c{Method}\\
    \midrule
    \cmark&
    \xmark&
    Input Img.&
    \inputImage&
    RGB Img. w/o \ac{MB}&
    \noMBImage&
    \mbRemoval&
    Li et al.\cite{li2023shiftnet}\\ 
    \cmark&
    \xmark&
    Input Img.&
    \inputImage&
    RGB Img. w/o \ac{DoF}&
    \noDoFImage&
    \dofRemoval&
    Ruan et al. \cite{ruan2022learning}\\ 
    \cmark&
    \xmark&
    Input Img.&
    \inputImage&
    RGB Img. w/o noise&
    \noNoiseImage&
    \noiseRemoval&
    Zamir et al. \cite{zamir2022restormer}\\ 
    \midrule
    \cmark&
    \cmark&
    Sharp Img.&
    \noMBImage, \mbParams, \flowImage&
    RGB Img. w/ \ac{MB}&
    \mbImage&
    \mbSynthesis&
    Potmesil and Chakravarty \cite{potmesil1983modeling}\\ 
    \cmark&
    \cmark&
    Sharp Img.&
    \noDoFImage, \dofParams, \depthImage&
    RGB Img. w/ \ac{DoF}&
    \dofImage&
    \dofSynthesis&
    Potmesil and Chakravarty \cite{potmesil1982synthetic}\\ 
    \cmark&
    \cmark&
    Noise-free Img.&
    \noNoiseImage, \noiseParams&
    RGB Img. w/ noise&
    \noiseImage&
    \noiseSynthesis&
    Trivial\\ 
    \midrule
    \cmark&
    \xmark&
    Input Img.&
    \inputImage&
    Flow map&
    \flowImage&
    \flowAnalysis&
    Ilg et al. \cite{ilg2017flownet}\\ 
    \cmark&
    \xmark&
    Input Img.&
    \inputImage&
    Depth map&
    \depthImage&
    \depthAnalysis&
    Ranftl et al. \cite{ranftl2021vision, ranftl2022midas}\\ 
    \bottomrule
    \end{tabular}
\end{table*}

\paragraph{Synthesis}
We can then make use of off-the-shelf, black box synthesis methods
$\mbSynthesis(\cdot,\mbParams)$,
$\dofSynthesis(\cdot,\dofParams)$ and
$\noiseSynthesis(\cdot,\noiseParams)$, that take physical parameters \mbParams, \dofParams and \noiseParams as arguments that we wish to discover.
For \ac{MB}, this is $\mbParams \in [0,1]$, the proportion of the frame interval over which the camera exposes the sensor.
For \ac{DoF}, \dofParams is the \remove{radius} \add{\ac{std}} of a gaussian blur kernel, as a function of depth.
We model this function as a quadratic \add{$\delta = G(Z) = AZ^2+BZ+C$}, meaning 3 real-valued parameters \add{$(A, B, C)$} are estimated.
For noise \noiseParams, it is the parameters of a noise generation model, such as the one outlined below.

How some of these parameters would act on an image depends on the space-time geometry underlying that image:
\ac{MB} depends on motion \flowImage (an optical flow map), \ac{DoF} would depend on the distance to the focal plane, according to \depthImage (a depth map) of each pixel and even noise depends on the pixel intensity itself.

If we have the parameters (\mbParams, \dofParams, and \noiseParams) as well as the space-time geometry (\depthImage and \flowImage) we can re-synthesize \ac{MB}, \ac{DoF} and noise, respectively, from the clean targets \noDoFImage, \noMBImage and \noNoiseImage, as in

\begin{align}
\dofImage(\dofParams)
&=
\dofSynthesis(\noDoFImage, \dofParams, \depthImage)
\text{,}\\
\mbImage(\mbParams)
&=
\mbSynthesis(\noMBImage, \mbParams, \flowImage)
\quad
\text{and}\\
\noiseImage(\noiseParams)
&=
\noiseSynthesis(\noNoiseImage, \noiseParams)
.
\end{align}

We assume \ac{DoF} and \ac{MB} to occur in this order and noise to happen last, which avoids incorrectly applying blur to the noise.

\paragraph{Analysis}
Fortunately, we can use a third pair of off-the-shelf components to extract scene parameters as well, as in 
\[
\depthImage=
\depthAnalysis(\inputImage)
\quad
\text{and}
\quad
\flowImage=
\flowAnalysis(\inputImage)
.
\]
Please again see \refTab{Ops} for the off-the-shelf-methods used. 
\new{Please note that the MB removal \cite{li2023shiftnet} and flow map estimation \cite{ilg2017flownet} requires frames adjacent to the current input frame as input. Since we deal with video pass-through mixed-reality, we assume adjacent frames are implicitly available.}

\paragraph{Noise}
Our noise model uses pyramid-based texture synthesis, similar to Portilla and Simoncelli \cite{portilla2000parametric} but greatly simplified as we only model sensor noise.
First we estimate the noise in the input image: $\inputImage - \noNoiseImage$.
We construct a Laplacian pyramid from this noise map, and estimate local \acp{std} of each noise level in each color channel, using the method outlined in Walton et al. \cite{walton2021beyond}.
We then fit a linear model mapping luminance values of the denoised image \noNoiseImage to these statistics at each pyramid level, via least-squares fitting. 
In practice we sample a subgrid of the pixels of stride 32 for greater efficiency. Noise for composited content is synthesised by using its luminance to estimate \acp{std} for each pyramid level, using the linear model.
These weight the levels of a Laplacian pyramid of a random Gaussian noise image.
Reconstructing from this produces the output noise, which is added to the virtual content.

\paragraph{Optimization}

For \ac{DoF} and \ac{MB} we first define the loss to optimize for \dofParams and \mbParams followed by optimization details.

\mymath{\aggregate}{\mathbb E}
\begin{align}
\dofLoss(\dofParams)
&=
\aggregate\left[
|\inputImage-\dofImage(\dofParams)|^2
\right]
\quad
\text{and}\\\
\mbLoss(\mbParams)
&=
\aggregate\left[
|\inputImage-\mbImage(\mbParams)|^2
\right]
\qquad\
\end{align}
Where \aggregate is the mean across all pixels. 
%
%
%
Our \ac{DoF} optimization proceeds in three stages. First, we exhaustively search \remove{(100 steps)} a range of gaussian blur \add{\acp{std} between} \remove{radius} $[0, 10]$ \add{(100 steps)} \remove{in pixels}, repeatedly applying it to de-blurred images and computing the error $|\inputImage - \noDoFImage(\dofParams)|^2$, recovering the optimal per-pixel \add{\ac{std}} \remove{radius}. 
Secondly, we filter out homogeneous non-textured regions which do not contribute meaningful blur information.
Finally, a polynomial degree-2 model $\dofParams = G(Z)$ is fit using an L-BFGS optimizer, mapping per-pixel depth to blur \add{\acp{std}} \remove{radius} for a given image.

Similar to \ac{DoF}, we optimize exposure \mbParams for MB as a function of optical flow estimated by running an exhaustive search \remove{(10 steps)} over a range of exposures $[0, 1]$ \add{(10 steps)}. 
We apply motion blur repeatedly with each exposure on the de-blurred image and compare with input image $|\inputImage - \noMBImage(\mbParams)|^2$. 
The exposure \mbParams is a global property for a given image; a function of the exposure settings of the camera and the optical flow estimated.




\new{The loss function used for the optimization of both \ac{DoF} and \ac{MB} parameters are typical $L_2$-norms.}
Where \noiseSynthesis, \mbSynthesis, \dofSynthesis are differentiable, optimization can be driven by gradient descent.
While the thin-lens model can be used for DoF optimization as proposed in \cite{chao2023imagestack}, we found our model provides better \new{performance and} flexibility with respect to the depth recovered from off-the-shelf NN methods, as it requires no prior optical information about the lens.

Differentiable proxies of the actual real-time method used at deployment could also be employed here.
\new{Fundamentally our system} is agnostic to the optimization method used as long as we can recover \noiseParams, \mbParams, and \dofParams for the \distortions models for deployment.

\paragraph{Speed}

We have implemented our optimization pipeline in a Python renderer. On average for an image of resolution $512\times$512, it takes \qty{36.3} millisec. to fit the noise generator, \qty{0.235} sec. to recover \mbParams for MB, and \qty{0.895} sec. to recover \dofParams for DoF. 

While the optimization is not real-time for all parameters, we note that each optimization needs to run (individually) only when the camera parameters 
or scene composition 
change. 
Most AR devices such as HMDs and mobile phones do not exhibit or desire rapid change in focus especially in the AR use case.
In the case of lighting changes on a device with automatic exposure and ISO settings, a conservative upper bound of 1.5 sec. on execution time (including off-the-shelf NNs) allows for fast enough adaptation that does not impair the user experience.
Typically optimization can be run once for each device at start-up (during boundary or hand-tracking calibration) and can be re-run regularly to update the parameters, as shown in the supplemental video.

\mysubsection{Testing}{Testing}

Deploying this model derived on real content is straightforward for virtual content (lower part of \refFig{Overview}).
We obtain the depth \depthImage, motion \flowImage and the virtual RGB image from deferred rendering.
To apply \distortions, we use the shutter time and circle of confusion, that will be mapped to suitable image blurs using  black-box-methods \mbSynthesis and \dofSynthesis robustly at real-time rates.

Most rendered examples shown here use differential rendering \cite{debevec1998rendering, kronader2015rendering} to improve realism. We can also apply our \distortions to the differential rendering buffers
and have appropriate effects on virtual shadows as well.

\newcommand{\method}[1]{\texttt{\textbf{#1}}}
\newcommand{\methodNaive}[1]{\textcolor{colorE}{\texttt{\textbf{#1}}}}
\newcommand{\methodOkumura}[1]{\textcolor{colorB}{\texttt{\textbf{#1}}}}
\newcommand{\methodOkumurapp}[1]{\textcolor{colorC}{\texttt{\textbf{#1}}}}
\newcommand{\methodMandl}[1]{\textcolor{colorD}{\texttt{\textbf{#1}}}}
\newcommand{\methodOurs}[1]{\textcolor{colorA}{\texttt{\textbf{#1}}}}

\mysection{Results}{Results}

We report qualitative results (\refSec{Qualitative}), analyze if we actually recover the particularities of specific cameras \new{on real scenes} (\refSec{Correctness}), compare to alternative approaches (\refSec{Comparison}), run a user study (\refSec{UserStudy}), \new{validate our system on synthetic content (\refSec{Validation})} and conclude with a demonstration of integration into a popular commercial MR system (\refSec{Demo}). 
In all results, na\"ive or \methodNaive{Naive} control is a straightforward alpha compositing of the original frame with rendered virtual objects using rendered objects' mask.
A list of cameras and lenses used for all \new{real scene} results are provided in the supplemental PDF.

\mysubsection{Qualitative}{Qualitative}

Qualitative results are shown in \refFig{Result} as well as in the accompanying supplemental video.
We used animal figurines together with their digital twins, animated virtual objects (butterfly, spaceship, pokemon) and dynamic real and virtual balls.

\refFig{Result}a shows a static virtual zebra placed at an out-of-focus depth.
Our method adds the correct blur, similar to the real blur on the real tiger in the background.
\refFig{Result}b shows bouncing balls with substantial motion blur and depth of field (note the tablecloth blurs into the back wall).
Our approach reproduces both. The supplemental video shows how a resting ball has correct depth-dependent blur and speed-dependent directional blur.  

\refFig{Result}c processes an input with significant noise, while retaining \ac{DoF} in the input and adding \ac{MB} to the output (butterfly in foreground).
The insets compare real and virtual noise that match, also in the presence of blur. \refFig{Result},d,e and f shows noise in combination with stronger motion from falling balls, swinging toy pendulum, and stronger defocus of closeup real and virtual objects.

\mycfigure{Correctness}{
Testing correctness of our approach by estimating camera parameters for different conditions.
The first three columns are three different input images.
In each column, this input is subject to increasingly strong distortion.
The distortion is \ac{MB} with falling balls in the first row (varying exposure), noise in the second row (varying exposure), and \ac{DoF} in the third row (varying focus distance).
Now, the last three columns are results of our algorithm, adding the right distortion, at the right extent to the same virtual content augmenting the input.
We see that the virtual object looks different for different levels of distortion.
We also see that the level of distortion is consistent with the input when composed.
Please see the supplemental PDF for a discussion on the parameters recovered and the supplemental video to see the distortions in motion.}

\mysubsection{Correctness}{Correctness}
We have seen that our approach can add faithful \distortions to virtual content, but do those distortions actually match those of the camera used.                                                                    
To better understand this relation, we have applied our approach to different levels of \distortions as well as different cameras. In \refFig{Correctness} we have applied our approach to the same scene, using the same camera, but at different settings.
Please see the caption of the figure for a discussion.
In \refFig{Cameras} we captured the same scene with multiple cameras and apply our method.
If our method works, it should extract different, correct parameter settings for each camera.
Please see the caption of \refFig{Cameras} for specific cameras used, and discussion of results.

\myfigure{Cameras}{Applying our method to augment very similar scenes with the same virtual objects captured with entirely different cameras.
We show original input images (top) for each augmented image (bottom).
We see that, while distortions differ, our methods consistently transfers them all with plausible settings to the virtual object.
Note that the scene framing differs, as the aspect and optics of these cameras differ.
Also note that the difference in color reproduction (radiometric calibration) are -- besides the chromatic variance of the noise-- not the aim of this work.}

\mysubsection{Comparison}{Comparison}

\paragraph{Baselines}
We look into three baselines, which all require known calibration targets, while our requires none (\refTab{baselines}).

The first baseline is \methodOkumura{OkumuraEtAl} \cite{okumura2006augmented}, which fits a motion-dependent Gaussian to an image of a known calibration target. 
Besides requiring this calibration target, which we do not, modeling \distortions with a single global elliptical blur kernel cannot handle more complex scenes where pixels have diversity of motion and depth.
We hence use a second baseline, of our own making, \methodOkumurapp{OkumuraEtAl++}, that tunes our parameters, but based on capturing their known calibration pattern.

The third and final baseline is \methodMandl{MandlEtAl} \cite{mandl2021neural}, which also uses prior calibration.
This method combines \distortions and radiometric compensation into a system of neural networks trained on images of the calibration target.
We do not perform radiometric compensation in our work and hence only use the ``LensNet'' part of their work, but apply it to scenes with relatively correct radiometrics, or at least radiometrics that are of similar quality level for all methods.

\begin{table}[ht]
\caption{Feature matrix for the baselines compared to our technique.}
\label{tab:baselines}
\centering
\begin{tabular}{rccccc}
\toprule
\multicolumn1c{Method}&
\multicolumn1c{Blind}&
\multicolumn1c{\ac{MB}}&
\multicolumn1c{\ac{DoF}}&
\multicolumn1c{Noise}\\
\midrule
\methodOkumura{OkumuraEtAl} \cite{okumura2006augmented}&
\xmark&
(\cmark)&
(\cmark)&
\xmark
\\
\methodOkumurapp{OkumuraEtAl++} \cite{okumura2006augmented}&
\xmark&
\cmark&
\cmark&
\xmark\\
\methodMandl{MandlEtAl} \cite{mandl2021neural}&
\xmark&
\xmark&
\cmark&
\cmark\\
\methodOurs{Ours}&
\cmark&
\cmark&
\cmark&
\cmark\\
\bottomrule
\end{tabular}
\end{table}

\paragraph{Outcome}

The outcome of this comparison is shown in \refFig{Comparison}. 
We note that, in contrast to the baselines, we are able to reproduce the desired \ac{MB} and \ac{DoF} effects without the need for marker tracking or prior camera calibration. In fact because of the distortion in images due to the blur effects, pose estimation with markers suffers resulting in further degradation in quality of results.

\newcommand{\task}[1]{\textsc{#1}}

\myfigure{Study}{\SP{Updated} Results of our \task{Identification} and our \task{Consistency} study on the top and bottom, respectively.}

\begin{table}[h]
    \caption{\SP{Updated} User study outcome. P-values for statistical tests relative to \methodOurs{Ours} are shown, in bold where significant ($p < 0.01$).}
    \label{tab:UserStudy}
    \centering
    \begin{tabular}{rrrrr}
    \toprule
    Method&
    \task{Ident.}&
    p=&
    \task{Cons.}&
    p=\\
    \midrule
    \methodOkumura{OkumuraEtAl}  \cite{okumura2006augmented}&
    47.4\,\%&
    0.066&
    55.3\,\%&
    0.177
    \\
    \methodOkumurapp{OkumuraEtAl++} \cite{okumura2006augmented}&
    49.3\,\%&
    0.134&
    13.2\,\%&
    \textbf{3e-15}
    \\
    \methodMandl{MandlEtAl} \cite{mandl2021neural}&
    19.1\,\%&
    \textbf{4e-12}&
    57.9\,\%&
    0.342
    \\
    \methodNaive{Naive}&
    14.5\,\%&
    \textbf{3e-15}&
    32.5\,\%&
    \textbf{2e-6}
    \\
    \methodOurs{Ours}&
    \textbf{57.9\,\%}&
    -&
    \textbf{64.0\,\%}&
    -
    \\
    \bottomrule
    \end{tabular}
\end{table}

\mysubsection{User study}{UserStudy}
We conducted a user study to assess the effectiveness of our system. 
\remove{The study was approved by ANON Ethics Board.} 
\add{This study involved human subjects in its research. Approval of all ethical and experimental procedures and protocols was granted by UCL Computer Science Research Ethics Committee (UCL/CSREC/R/16).} 

\paragraph{Setup}
Stimuli were still images that contain \ac{MB} and \ac{DoF} blur effects with augmentations, similar to those seen in \refFig{Comparison}, with a single virtual object (animal figurine), where consistency was achieved by either, one of our three baselines \methodOkumura{OkumuraEtAl}, \methodOkumurapp{OkumuraEtAl++}, \methodMandl{MandlEtAl},  by the \methodNaive{Naive} (na\"ive) control or by \methodOurs{Ours}.
\add{All order of methods and scenes were randomized.}
Users performed two tasks:

In Task 1: (\task{Identification}), subjects viewed an image containing animal figurines and at most one virtual object on top of a marker.
They were asked: ``Is there a virtual object in this image?''. 
Our goal was to find which technique produces results more often considered real by subjects unaware of the augmentations.
All techniques were used with the same figurines and position for consistency on lighting and point of view. 
We did not include scenes with all real figurines, but mixed real with virtual ones (see \refFig{Comparison}).
Our concern was that including real-only scenes would cause users to focus on slight shading differences between real and virtual objects rather than motion and defocus blur effects.

In Task 2: (\task{Consistency}), subjects watched an 8 second video clip of a camera moving around a marker with a virtual Zebra or Tiger rendered with all of the evaluated techniques.
This allowed a proper assessment of the synthesized motion blur, as it relates to the camera movement and isolated frames could be misleading.
The question asked was ``Is blur on the virtual animal figurine consistent with the background?''. 

\paragraph{Outcome}

Outcome of this study is shown in \refFig{Study} and \refTab{UserStudy}.
We recruited a total of \qty{19} participants for the study\remove{via Google forms}. 
\add{We did not control for demographics as we did not expect any effect on low-level vision \cite{Shaqiri2018}. 
Participants were recruited by email to authors' colleagues and the study was conducted over anonymized Google forms. The demographics consisted of students and researchers from higher education institutions in the US and Europe. In general, the subjects had research background in HCI, computer vision \& graphics, and/or psychology.}

For the \task{Identification} task, we report the percentage of trials in which participants failed to identify virtual objects.
For \task{Consistency} we report the percentage where participants reported blur on virtual figurine being consistent with the real figurines.
Thus for both tasks, a higher result is better.

In both trials our method achieved the best results overall. 
We assess the significance of this result, comparing our results to those of the other approaches using a two-proportion z-test \cite{statspython2023}.
We found that our method produced significantly better results than \methodMandl{MandlEtAl} on the \task{Identification} task and \methodOkumurapp{OkumuraEtAl++} on the \task{Consistency}, and significantly  outperformed the \methodNaive{Naive} approach in all cases. For other cases although results were not statistically significant, we can conclude our method offers similar performance to state-of-the-art approaches, despite not requiring markers or pre-calibration.

\myfigure{Validation_params}{Recovered parameters using \task{Optimization} and \task{FullPipeline} on the top and bottom, respectively for \ac{MB} (left) and \ac{DoF} (right). For \ac{MB} ground truth parameters are shown in \emph{blue} and recovered parameters are shown in \emph{orange}. For \ac{DoF} ground truth models $G(Z)$ are shown with \emph{solid} lines and recovered models $\hat{G}(Z)$ are shown with \emph{dotted} lines. The colors denote different models of \ac{DoF}.
We observe good correlation between ground truth and recovered parameters in both experiments.
}

\myfigure{Validation}{Validation frames for \task{Optimization} and \task{FullPipeline} on the top 2 rows and bottom 2 rows, respectively for \ac{MB} (first and third row) and \ac{DoF} (second and fourth row). 
In the top 2 rows, we show (L-R) sharp distortion-free image, GT distorted image, image distorted with recovered parameters and the mean squared error map between GT and distorted images.
In the bottom 2 rows, we show (L-R) undistorted image as outcome of off-the-shelf distortion removal methods (\refTab{Ops}), GT distorted image, image distorted with recovered parameters and the mean squared error map between GT and distorted images.
We observe good recovery of parameters verified by low error on the error maps between ground truth and distorted images.
}

\myfigure{Validation_noise}{Noise model validation using \task{Optimization} and \task{FullPipeline} on the top and bottom, respectively. 
We show (L-R) ground truth noisy image, our estimated noisy image, instance of the residual noise, GT noise which is same as the residual noise in \task{Optimization} and recovered noise.
Please observe the residual noise for \task{FullPipeline} has spatial structure corresponding to discontinuities in the image which indicates that the denoiser smooths high frequency details during denoising.
Regardless, our estimated noise model generates noise in distribution of ground truth noise. 
}


\mysubsection{Validation}{Validation}
In previous sections we demonstrated effectiveness of our system in creating consistent composites on real scenes. Furthermore, to validate how well our system works, we evaluate our pipeline on synthetically rendered video frames. First, we generate synthetic ground truth images by rendering a virtual scene using Blender and distort them with \emph{known} \distortions model parameters giving us ground truth. Next, we evaluate our system to recover the model parameters and compare it with the ground truth.

\paragraph{Data}

For \ac{MB} we evaluated our system using \qty{10} models with increasing exposure time. To evaluate \ac{DoF} we used \qty{5} models with different focal plane and shallowness of depth-of-field. To evaluate noise, we use hetroscedastic Gaussian model, with $\sigma(I) = k_{readout} + k_{shot} \sqrt{\mu(I)}$ as defined in \cite{klein2009simulating, maleky2022noise2noiseflow}, which is widely believed to most closely resemble sensor noise profile of real cameras. We used \qty{10} different random parameters for $k_{readout}$ and $k_{shot}$. 
In total, we rendered \qty{100} frames of a scene and distorted each frame with all distortion parameters. 

\paragraph{Experiments}
As an advantage of using synthetic images, we automatically obtain ground truth (GT) distorted-undistorted image pairs and scene geometry using deferred rendering. Hence, in our first experiment \task{Optimization}, we test optimization efficiency by using perfect distortion-free images, depth and flow maps to recover \dofParams, \mbParams, and \noiseParams from synthetically distorted images. 

In another experiment \task{FullPipeline}, we test our full pipeline by running off-the-shelf methods from \refTab{Ops} on synthetically distorted images, followed by running the optimization to recover distortion model.
\task{FullPipeline} evaluates our system, while \task{Optimization} shows how well it could perform if perfect distortion-free images and scene geometry were recovered or available.

\paragraph{Outcome}

\refFig{Validation_params} (top row) shows comparison of the estimated model with the ground truth for MB and DoF as a result of \task{Optimization}. 
For \ac{MB} we average the recovered parameters over the 100 frames in the shown plot. 
We observe good correlation between GT and estimated parameters which confirm that \task{Optimization} is able to recover the \ac{MB} parameters well.
For \ac{DoF} we show the recovered models for 1 frame (shown in \refFig{Validation}).
Similar to \ac{MB}, we observe good correlation between GT model and our recovered model as a result of \task{Optimization} for \ac{DoF} as well.

\refFig{Validation_params} (bottom row) shows comparison of the estimated model with the ground truth for MB and DoF as a result of \task{FullPipeline}. 
In both \ac{MB} and \ac{DoF} case, we see similar behavior as \task{Optimization} with some degradation in accuracy.

For \ac{MB} our estimated exposure parameters follow the same trend as the ground truth.
In case of \ac{DoF}, the estimated models maintain the focal region and extent of depth-of-field.
Our recovered \ac{DoF} models are plausible but inexact, such as red model in \task{Optimization} (\refFig{Validation_params}), which can be due to insufficient data points at specific depths.
In \task{FullPipeline}, we attribute the deviation from GT to inaccuracies in recovered sharp images and scene geometry compounded by insufficient data points.
Visually the estimated frames are very similar to GT frames for \ac{MB} and \ac{DoF} as shown in \refFig{Validation}.

We show outcome of noise model validation in \refFig{Validation_noise}.
Please see the caption for a discussion of the estimated noise model validating our noise estimation works well.

\myfigure{Demo}{Captured frames of our real-time demo, where objects move and a user moves their head.
For discussion, see \refSec{Demo}.}

\mysubsection{Real-time Demo}{Demo}
The results in the previous sections \add{(4.1-4.5)} used a \remove{non-realtime} Python renderer \add{whose speed is reported in Section 3.1}.  
We have also implemented the \distortions in a Unity application that demonstrates that once the \distortions parameters are obtained (within a few seconds), they can be applied in real-time.
We \remove{will} report these for a Unity demo that has industry-level implementations of \distortions available.
Using our parameters there shows that they are interpretable and applicable, and gives our final performance numbers.

The demo runs at average \qty{45}\Hz~at 2064$\times$2208 pixels per-eye resolution.
Please see the supplemental video for a capture.

We chose a Meta Quest 3 as the device for our demos, as it is currently the most popular video see-through AR HMD in the market (Qualcomm Snapdragon XR2 Gen 2, CPU: Octa-core Kryo, GPU: Adreno 740, 8GB RAM).
The prototype was implemented using Unity's Universal Rendering Pipeline (version 2023.2.2f1), using OpenXR, AR Foundation and the Meta OpenXR packages for Passthrough support.
Depth of field, Motion-Blur and Film-Grain were applied as post-processing effects \footnote{Adapted to support alpha compositing on version 2023.2.2f1.} using parameters estimated by our method using a sample capture under fixed lighting conditions.

\mycfigure{Result}{Typical results produced by our approach.
We always show a pair of \methodNaive{Naive} (na\"ive) compositing (columns 1,3) and \methodOurs{Ours} (columns 2,4).
Below each pair, we show insets from both methods.
Please see \refSec{Results} for a discussion.
Please also see the supplemental video for animated version of those figures and the supplemental PDF for additional results.}

\mycfigure{Comparison}{Comparison between different methods \textbf{(columns)} on different scenes \textbf{(rows)}.
Please note that all methods except \methodOurs{Ours} either have seen the marker and need it to be present (\methodOkumura{OkumuraEtAl} and \methodOkumurapp{OkumuraEtAl++}) or need previous calibration on a known object from the scene \methodMandl{MandlEtAl}.
Our methods does not use that marker, neither did it see the scene before.
We add a virtual Zebra figurine (rows 1,2,3), virtual Tiger figurine (rows 4,5) and a virtual Chimpanzee figurine (row 6) to the scene.
All other objects are real.
In terms of quality, all methods reproduce some form of blur that is consistent with the image.
\methodOkumura{OkumuraEtAl} adds blur, but it is not directional and does not depend on the location inside the scene, as seen in the last scenes with multiple objects.
\methodOkumurapp{OkumuraEtAl++}, our improved baseline, resolves this issue \add{but tended to produce exaggerated blur in some cases}.
\methodMandl{MandlEtAl} has more refined blur operations that fit the scene's blur situation the same as \methodOurs{Ours} does, but we did not use any previous calibration step.
Please see the supplemental video to observe and compare motion blur effects.}

\mysection{Conclusion}{Conclusion}

We have presented an approach to achieving consistency between virtual and real objects.
The key idea is to rely on modern computer vision to produce calibration data on the fly, and use that to tune existing real-time \distortions effects found in many \ac{MR} systems.
We have compared the approach to other methods, and show superior quality without markers, calibration targets or prior knowledge about the scene.
In a user study, comparing against marker-based and pre-calibrated methods, we outperformed previous methods in an identification task and in user preference.

\paragraph{Limitations}
We inherit the limitations of the trained \remove{blur and DoF} \add{distortion removal} models we employ, including reduced performance on data dissimilar to the training set of computer vision methods we use.


While we found the noise estimation to provide plausible results, our linear model will still predict noise on over-saturated regions where noise levels are above the cutoff point; if it finds a correlation between luminosity and noise.
A more complicated estimator could potentially address this, at the cost of reduced speed. 

\paragraph{Future work}

In future work, we would like to model the remaining pipeline elements that we did not yet consider, the ISP, the radiometrics, or other, unknown on-camera processing.
We are confident this is possible using the same approach, but also note that our approach could be combined with others that simulate parts of the pipeline explicitly.
Finally, we of course would like to further improve efficiency, accuracy and robustness of our tuning.

Reaching further, we would like to consider consistency in advanced modalities (stereo or light field displays, holograms) as well as mixed sensory experiences including augmentation of audio or haptics.

\maketitle
\section*{Supplemental Materials}
\label{sec:supplemental_materials}








\subsection*{Noise Model Details}

The noise model used in this paper proceeds in two main stages: first the model is fitted, based on an input image and noise map. Second, this model is used to synthesise noise for virtual content added to the scene.

\subsubsection*{Fitting}

The inputs to the fitting step are the input image \inputImage and the estimated denoised image \noNoiseImage produced by the denoising method.
From these we find an estimate for the ground truth noise $N_{GT} = \inputImage - \noNoiseImage$.

We first construct a Laplacian pyramid of $N_{GT}$. 
The examples in this paper make use of a 4-level Laplacian pyramid, which was sufficient as noise was typically of high frequency. 
The final level of the pyramid containing the means was ignored, as we assume the noise to have zero mean.

We compute local standard deviations for each level of this pyramid using the method of \cite{walton2021beyond}, to wit we use the standard formula:
\begin{equation}
    \sigma(X) = \mathbb{E}(X^2) - \mathbb{E}(X)^2
\end{equation}

We find local means by applying a blur $B$ to the pyramid level, so for the $l$th level of the pyramid $N_{GT,l}$:
\begin{equation}
    \sigma(N_{GT,l}) = B(N_{GT,l}^2) - B(N_{GT,l})^2
\end{equation}

In our case we choose $B$ to be a 9x9 Gaussian blur with $\sigma=4$ at the highest resolution pyramid level.
The value of $\sigma$ was halved at each level to provide the same effective pooling size at all resolutions.

At each pyramid level $l$ we then fit a linear mapping $L_l$ from the luminance at each pixel (i.e. Y channel of the image in YCbCr format) to the local standard deviations $\sigma(N_{GT,l})$.
Since we have standard deviations for the red, green and blue channels, this is a mapping $L_l: \mathbb{R} \rightarrow \mathbb{R}^3$.
This is fitted by least-squares linear regression, giving a 2x3 matrix encoding the model per pyramid level.
In practice we sample the images once every 32 pixels in each dimension to greatly reduce the number of samples our model has to fit to, meaning the least-squares fitting has to find the pseudo-inverse of a much smaller matrix.
This pixel skip is also halved as we move down each resolution level.

Our total model consists of a total of $4 \times 2 \times 3 = 24$ parameters.

\subsubsection*{Synthesis}

To synthesise noise to add to some noise-free virtual content $I_{\text{virt}}$, we first start from an image containing random normally-distributed noise $N_{\text{init}}$ over the RGB channels. 
A laplacian pyramid of this noise map is constructed, and each level is normalised to have local standard deviation of 1 at all locations.
This is achieved by finding the local standard deviations $\sigma$ as above, and simply dividing through:
\begin{equation}
    N_{\text{norm},l} = \frac{N_{\text{init},l}}{\sigma(N_{\text{init},l})}
\end{equation}

We find the luminance of the virtual content we wish to add the noise to $Y_{\text{virt}}$. 
For each pyramid level of the noise we input this to the corresponding linear mapping $L_l$ to obtain a map of standard deviations in red, green and blue.
We multiply these by the normalised noise pyramid levels $N_{\text{norm},l}$ to achieve the correct statistics at each pyramid level.
Finally, reconstructing from this Laplacian pyramid of weighted noise maps produces the synthesised noise, which can be added to the virtual content.

We note that though our current implementation is based in Pytorch, efficiency could be further improved by making use of graphics hardware accelerated MIP map generation to compute the pyramids as was done in Walton et al. \cite{walton2021beyond}.

\begin{table}[h]
    \caption{List of cameras and lenses used.}
    \label{tab:Devices}
    \centering
    \begin{tabular}{lr}
    \toprule
    Camera + Lens &
    Figure 
    \\
    \midrule
    Flir Chameleon 3 + 
    & Fig. 1, Fig. 2
    \\
    TV 16mm 1:1.4 & 
    Fig. 3 Row 1, 2, 
    Fig. 10 (a, c, d),
    \\
    & Fig. \ref{fig:Results_supp} (a, d) (supp)
    \\
    \midrule
    Sony Alpha 7III +
    & Fig. 3 Row 4, 5,
    \\
    Tamron 28-75 F/2.8 &
    Fig. 10 (b),
    Fig. 11,
    Fig. \ref{fig:Comparison_supp} (supp)
    \\
    \midrule
    Nikon D3100 + 
    & Fig. 3 Row 3
    \\
    Nikkor 55-200mm F/1.4-5.6 
    & Fig. 10 (e, f)
    \\
    \midrule
    Meta Quest 3 + on-device lenses
    & Fig. 9
     
    \\
    \bottomrule
    \end{tabular}
\end{table}



\begin{figure*}[htb]%
\centering\includegraphics*[width = \linewidth]{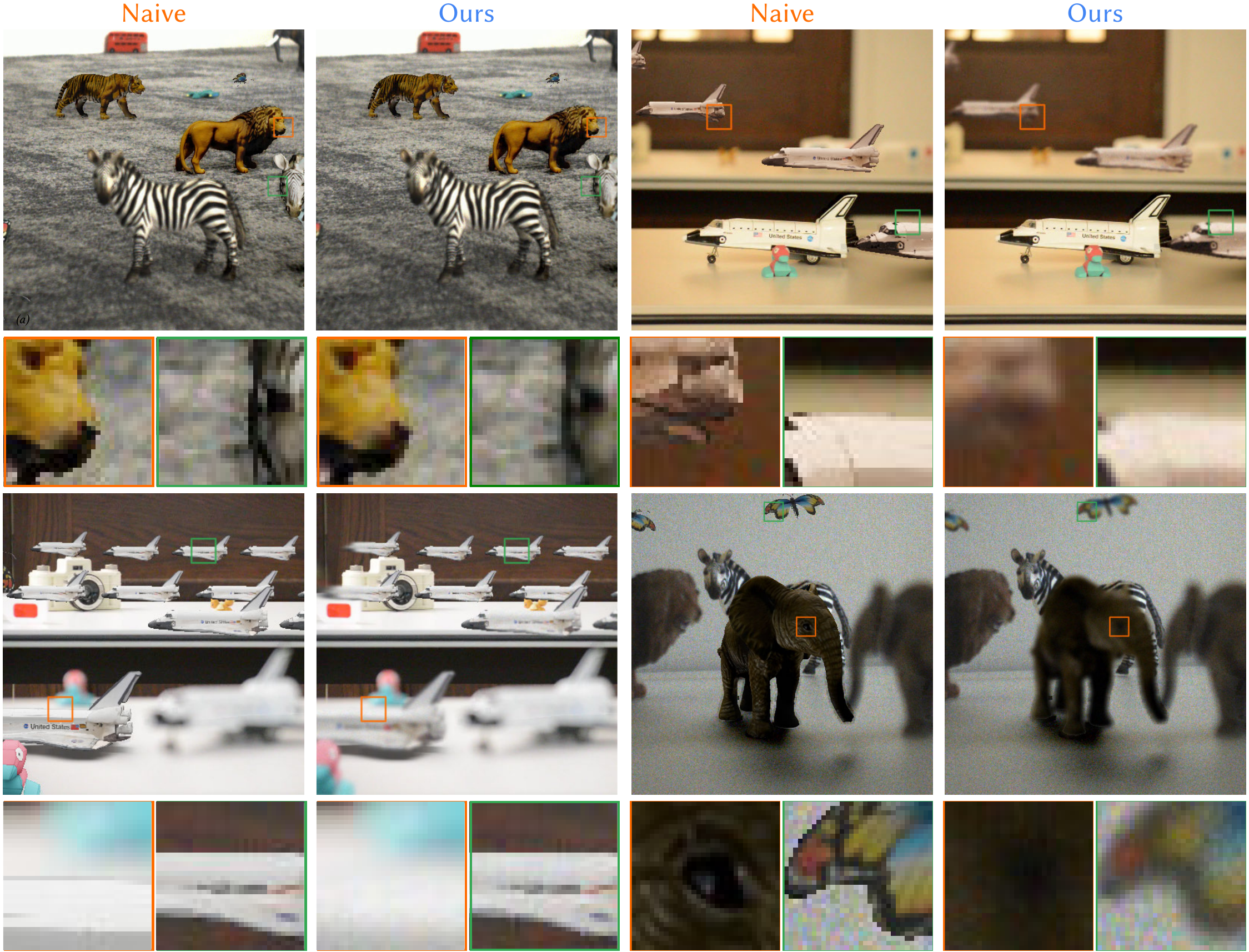}%
\vspace{-.1cm}%
\caption{Additional typical results produced by our approach.
We always show a pair of \methodNaive{Naive}(na\"ive) compositing (columns 1,3) and \methodOurs{Ours} (columns 2,4).
Below each pair, we show insets from both methods.}%
\vspace{-.2cm}%
\label{fig:Results_supp}%
\end{figure*}%

\begin{figure*}[htb]%
\centering\includegraphics*[width = \linewidth]{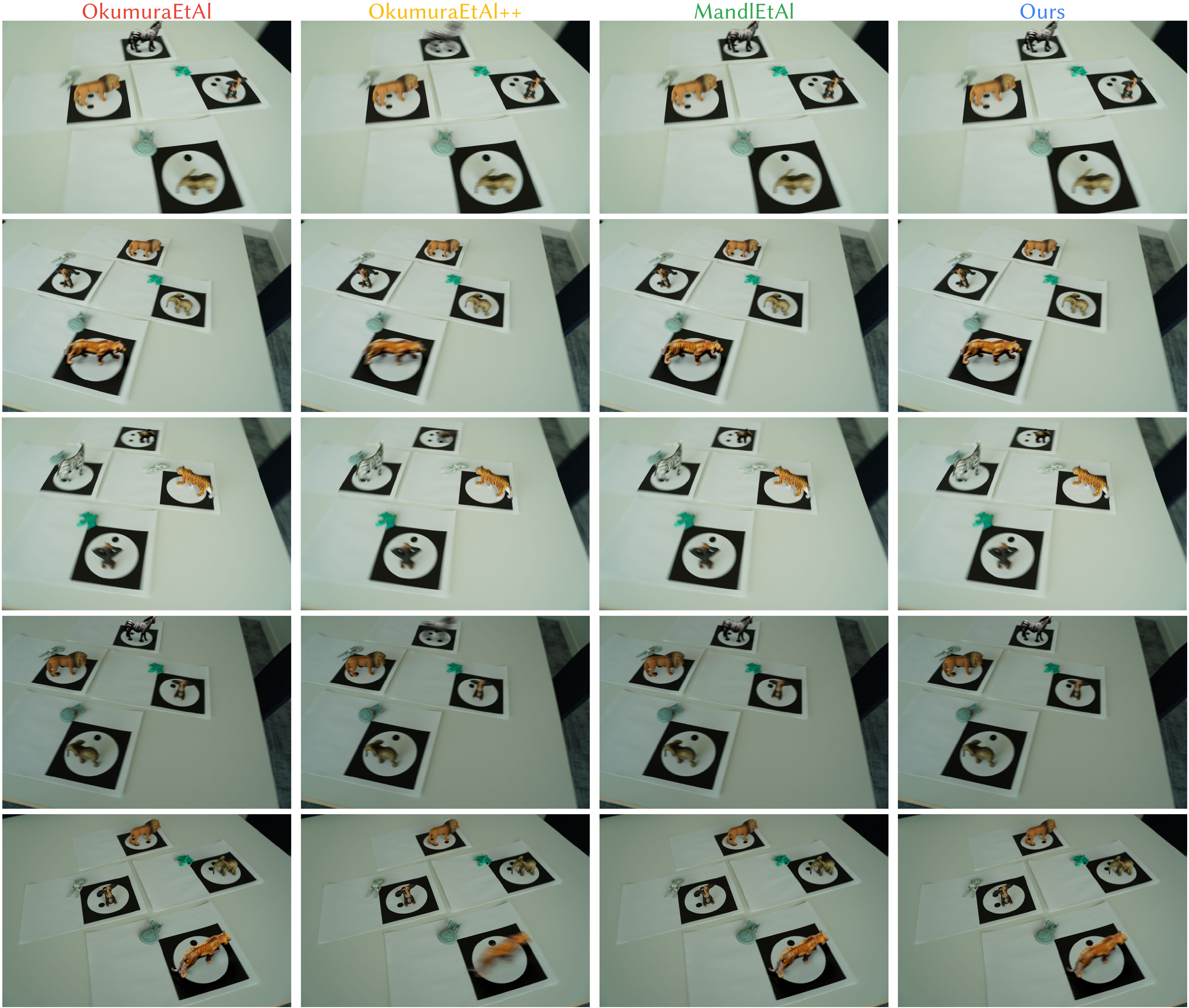}%
\vspace{-.1cm}%
\caption{Additional comparison between different methods \textbf{(columns)} on different scenes \textbf{(rows)}.
Please note that all methods except \methodOurs{Ours} either have seen the marker and need it to be present (\methodOkumura{OkumuraEtAl} and \methodOkumurapp{OkumuraEtAl++}) or need previous calibration on a known object from the scene \methodMandl{MandlEtAl}.
Our methods does not use that marker, neither did it see the scene before.
We add a virtual Zebra figurine (rows 1, 4), virtual Tiger figurine (rows 2, 5) and a virtual Elephant figurine (row 3) to the scene.
All other objects are real.}%
\vspace{-.2cm}%
\label{fig:Comparison_supp}%
\end{figure*}%

\begin{figure*}[htb]%
\centering\includegraphics*[width = \linewidth]{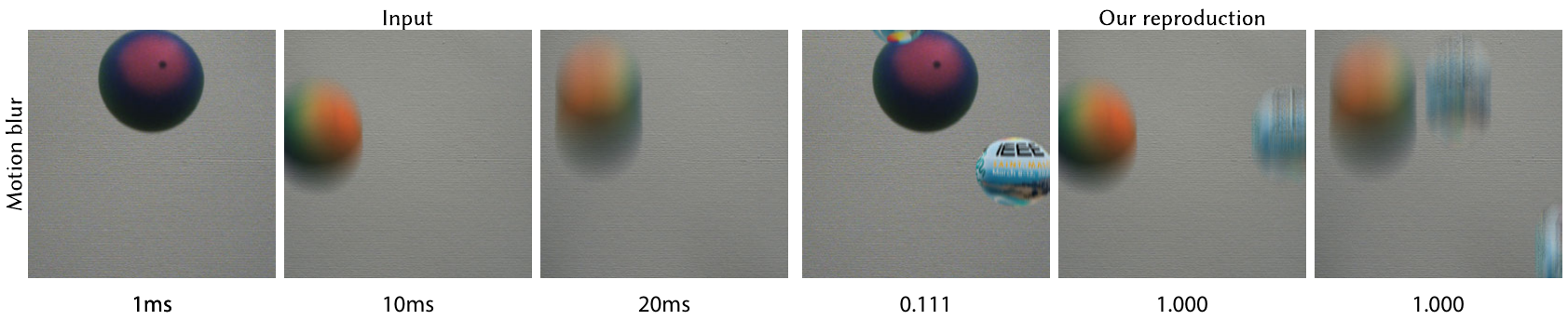}%
\vspace{-.1cm}%
\caption{Recovered parameters for \ac{MB}. On the left three columns we show images captured at varying exposure levels. This can be observed from varying level of motion blur in the real falling ball. On the right we have our reproduction of motion blur on virtual falling balls which is blurred with the parameters recovered from our optimization. The exposure parameters recovered are shown below each frame.}%
\vspace{-.2cm}%
\label{fig:Correctness_mb_parameters}%
\end{figure*}%

\begin{figure*}[htb]%
\centering\includegraphics*[width = \linewidth]{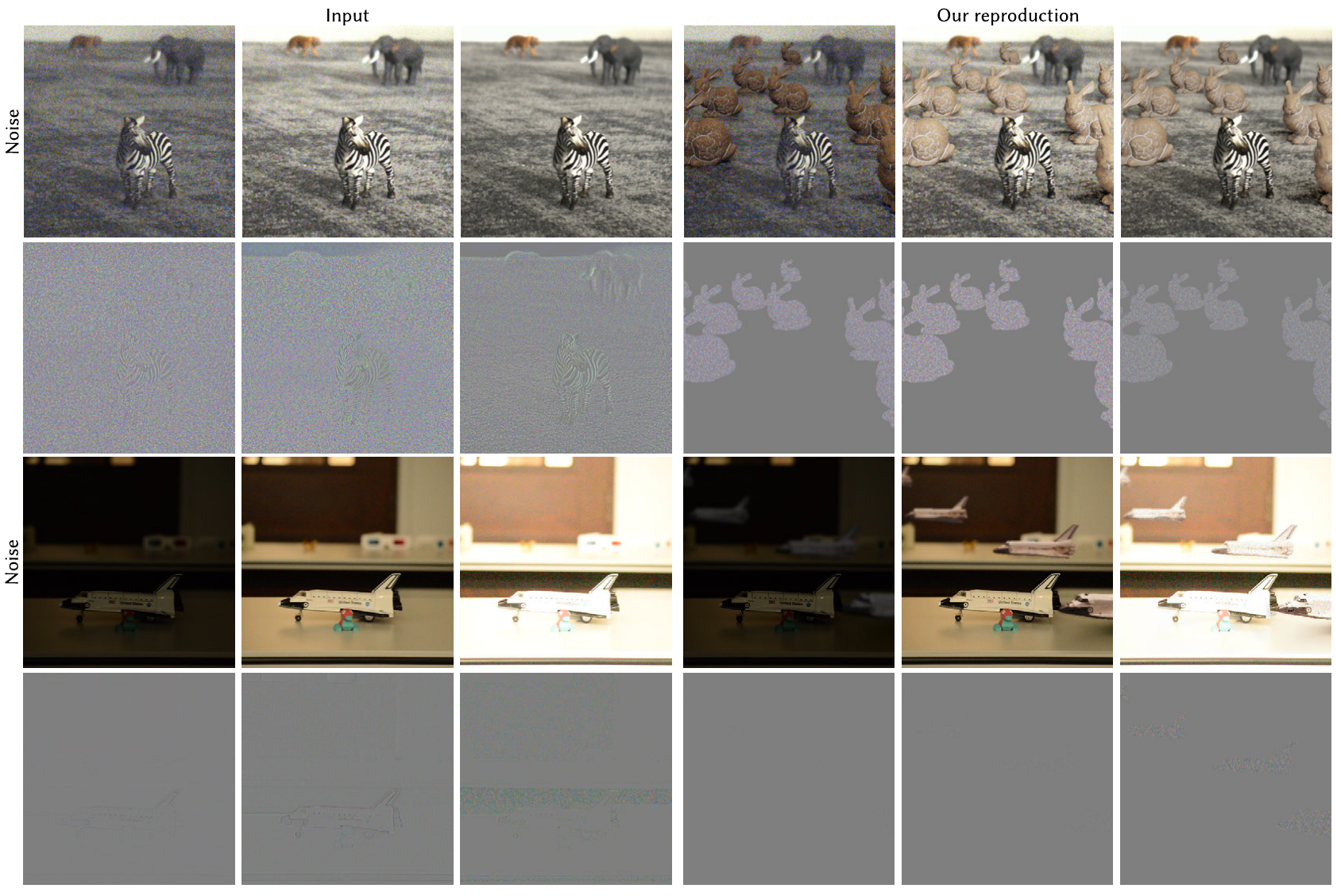}%
\vspace{-.1cm}%
\caption{Recovered parameters for noise. We show the residual noise from input images (left three columns, second and fourth row) and our generated noise (right three columns, second and fourth row) as a result of our optimization for the images (left three columns, first and third row). The generated noise are synthesized over the virtual composites (right three columns, first and third row) on the input images to make them consistent with overall noise in the image.}%
\vspace{-.2cm}%
\label{fig:Correctness_noise_parameters}%
\end{figure*}%

\begin{figure*}[htb]%
\centering\includegraphics*[width = \linewidth]{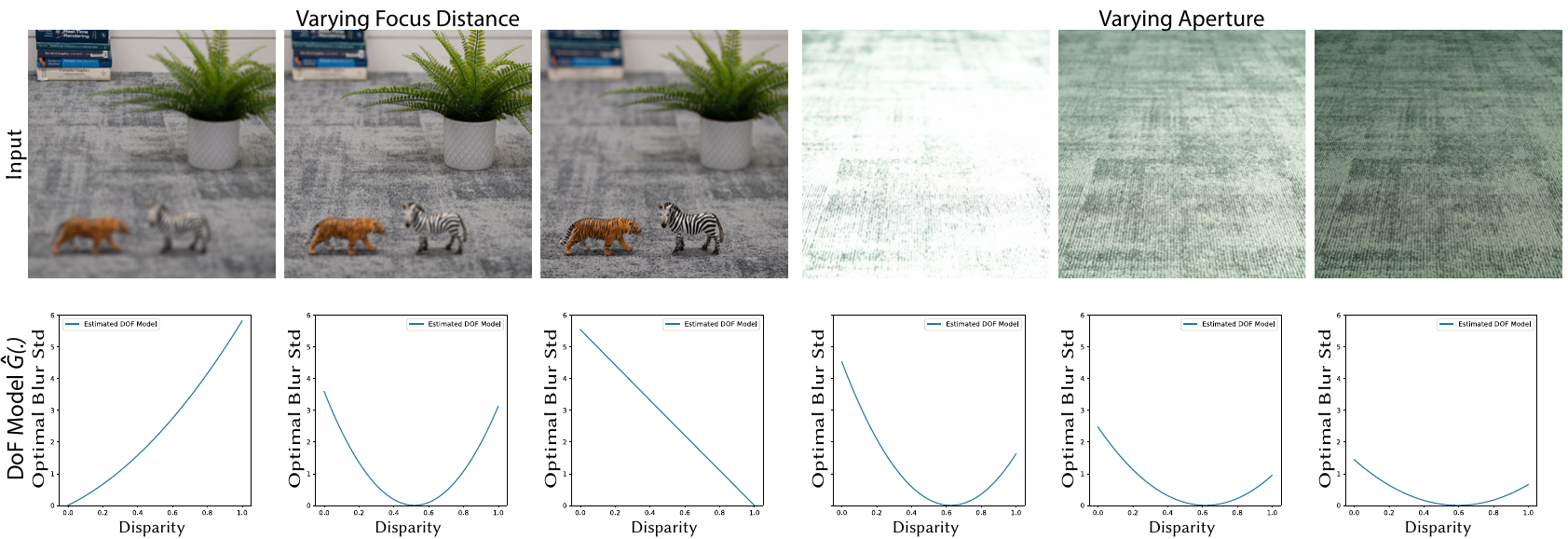}%
\vspace{-.1cm}%
\caption{Recovered parameters for \ac{DoF}. We show the \ac{DoF} model $G(.)$ (bottom row) recovered for different input images (top row) with varying parameters. In the left three columns we show input images with varying focus distance, focused at far, middle and near plane respectively. In the right three columns we show input images with varying aperture setting of f/2.8, f/5.6 and f/11.0 respectively. We notice our recovered model accurately predicts the focal plane as well as the shallowness of \ac{DoF} with decreasing (narrowing) aperture.}%
\vspace{-.2cm}%
\label{fig:Correctness_dof_parameters}%
\end{figure*}%

\subsection*{Additional Results and Comparison}

A list of cameras and lenses used for results in each figure is provided in Tab.~\ref{tab:Devices}.

We provide additional results similar to Fig. 10 and Fig. 11 of the main paper in Fig. \ref{fig:Results_supp} and Fig. \ref{fig:Comparison_supp}, respectively.

We also show the parameters recovered as a result of our optimization on the images in Fig. 3 of the main paper in Fig. \ref{fig:Correctness_mb_parameters}, Fig. \ref{fig:Correctness_noise_parameters}, and Fig. \ref{fig:Correctness_dof_parameters}.




\acknowledgments{%
	The authors wish to thank Sebastian Friston and David Swapp for early discussions, and David Mandl for providing the code for Neural Cameras~\cite{mandl2021neural}.
  This work was funded by the EPSRC/UKRI project EP/T01346X/1.%
}

\bibliographystyle{abbrv}

\bibliography{paper}

\end{document}